\documentclass{article}
\usepackage{ijcai23}

\usepackage{times}
\usepackage{soul}
\usepackage{url}
\usepackage[hidelinks]{hyperref}
\usepackage[utf8]{inputenc}
\usepackage[small]{caption}
\usepackage{graphicx}
\usepackage{amsmath}
\usepackage{amsthm}
\usepackage{booktabs}
\usepackage{algorithm}
\usepackage{algorithmic}
\usepackage[switch]{lineno}
\usepackage{multirow}
\usepackage{amssymb}
\usepackage{paralist}

\title{Decision Fusion Network with Perception Fine-tuning for Defect Classification}

\author{
Xiaoheng Jiang$^{1,3,4}$
\and
Shilong Tian$^2$\and
Zhiwen Zhu$^{1}$\and
Yang Lu$^{*1,3,4}$\and
Hao Liu$^{1,3,4}$\and \\
Li Chen$^{1,3,4}$\and
Shupan Li$^{1,3,4}$\and
Mingliang Xu$^{*1,3,4}$\and
\affiliations
$^1$School of Computer Science and Artificial Intelligence, Zhengzhou University, Zhengzhou, China\\
$^2$School of Computer Science and Engineering, Beihang University, Beijing, China\\
$^3$Engineering Research Center of Intelligent Swarm Systems, Ministry of Education, Zhengzhou, China\\
$^4$National Supercomputing Center in Zhengzhou, Zhengzhou, China
\emails
\{jiangxiaoheng, ieylu,  cli, iespli, iexumingliang\}@zzu.edu.cn,
tianshilong@buaa.edu.cn,
1277634792@qq.com,
HaoLiu1989@hotmail.com
}

\begin{document}

\maketitle

\begin{abstract}

Surface defect inspection is an important task in industrial inspection. Deep learning-based methods have demonstrated promising performance in this domain. Nevertheless, these methods still suffer from misjudgment when encountering challenges such as low-contrast defects and complex backgrounds. To overcome these issues, we present a decision fusion network (DFNet) that incorporates the semantic decision with the feature decision to strengthen the decision ability of the network. In particular, we introduce a decision fusion module (DFM) that extracts a semantic vector from the semantic decision branch and a feature vector for the feature decision branch and fuses them to make the final classification decision. In addition, we propose a perception fine-tuning module (PFM) that fine-tunes the foreground and background during the segmentation stage. PFM generates the semantic and feature outputs that are sent to the classification decision stage.  Furthermore, we present an inner-outer separation weight matrix to address the impact of label edge uncertainty during segmentation supervision. Our experimental results on the publicly available datasets including KolektorSDD2 (96.1\% AP) and Magnetic-tile-defect-datasets (94.6\% mAP) demonstrate the effectiveness of the proposed method.

\footnote{$^*$Corresponding author}
\end{abstract}

\section{Introduction}
Defect inspection plays a vital role in quality control during industrial production. The emergence of deep learning has significantly contributed to the development of defect inspection techniques. These techniques can be broadly classified into three categories: image classification-based methods~\cite{park2016ambiguous}, object detection-based method~\cite{he2019end,xie2020ffcnn}, and pixel segmentation-based methods~\cite{zavrtanik2021draem,jiang2022joint,zou2018deepcrack}. Image classification-based methods categorize images into defective and defect-free classes. Object detection-based methods treat defect detection as an object detection problem and localize defects using rectangular boxes. Pixel segmentation-based approaches treat defect detection as a task of segmenting pixels at a fine-grained level, thereby yielding more precise outcomes. 

We mainly focus on the defect classification task in our work. In the context of complex background, two primary challenges arise. The first challenge pertains to the issue of background interference. Some backgrounds are visually similar to defects, especially when the defects are small. On this condition, the classification network can wrongly judge the input images as defective ones. This predicament is exemplified in Figure 1(a).
The second challenge arises from the fact that the input images containing defects are low contrast. The defects are visually submerged by the background. On this condition, the classification network can wrongly judge the input images as defect-free ones. Figure 1(b) illustrates this kind of challenge. 
These two challenges commonly exist in many industrial scenes. Though many methods~\cite{zeng2022small,yang2021automatic,tao2020industrial,zou2018deepcrack} have been proposed to deal with these issues, they are still not satisfactorily addressed. 

\begin{figure}
	\centering
	\includegraphics[width = 0.48\textwidth]{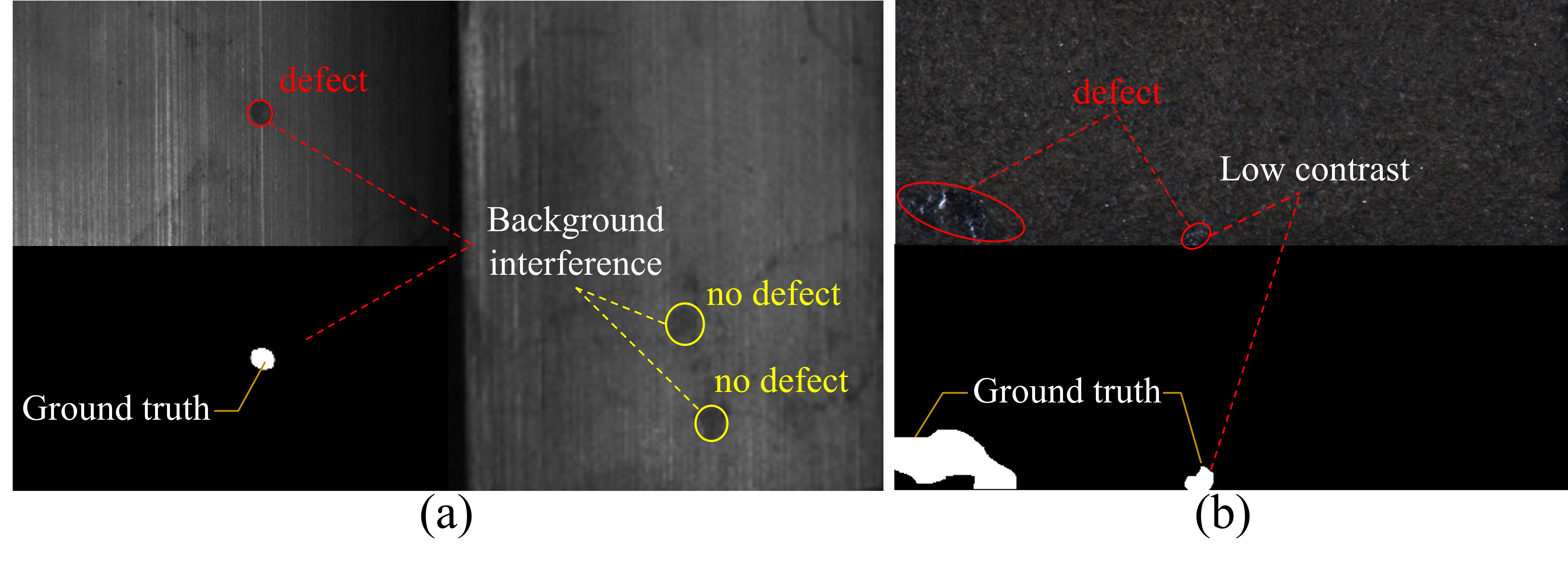}
	\caption{Some challenges: (a) shows the background interference in the Magnetic-Title dataset, and (b) shows the low-contrast defects in the KSDD2 dataset.}
\end{figure}

In this paper, we attempt to address the aforementioned issues from two perspectives. (1) We integrate the pixel-level segmentation information into the classification network to strengthen the ability of feature extraction about defects. (2) We adopt the decision fusion strategy to strengthen the robustness of the classification task. 
Based on the above analysis, we propose a two-stage  defect detection network that comprises a segmentation stage and a classification stage. The proposed network aims to enhance classification accuracy by integrating semantic defect information.

Specifically, we propose a decision fusion module (DFM) that leverages segmentation results to aid in decision-making. The DFM encodes and fuses the semantic information and CNN features, so as to make a comprehensive decision about whether there are defects in the input image. 
In addition, we introduce a perception fine-tuning module (PFM) in the first stage to generate semantic information and features about the defects. The PFM uses the initial segmentation information to make the features focus on the foreground and background, respectively. Subsequently, a refining module with dilated convolutions is applied to refine the foreground and background features. On the one hand, these features are used to generate the complementary segmentation maps, which are first combined with the initial segmentation map and are then sent to the classification decision stage. On the other hand, these features are combined with the initial features and are then sent to the classification decision stage.
Furthermore, to tackle the impact of the dilation-based pixel-level data augmentation during the segmentation learning, we introduce the inner-outer separation weight matrix (SWM). It preserves the weights of the precise areas in the augmented annotations while attenuating the weights of the dilated regions. As a result, it helps to obtain more precise segmentation results, which is important to the final defect classification task. 
In summary, the main contributions are as follows.

\begin{enumerate}
\item We propose a decision fusion network that jointly encodes the semantic information and CNN features of the defects, which helps to make a robust decision about the defects. 
\item We propose a perception fine-tuning module that can refine the foreground and background features and provide the corresponding complementary segmentation results, which enhances the ability of defect information extraction in complex backgrounds.
\item We introduce an inner-outer separation weight matrix when calculating the segmentation loss, which helps to improve the precision of the defect segmentation.
\end{enumerate}

The proposed approach has undergone evaluation on two publicly available datasets including KolektorSDD2~\cite{bovzivc2021mixed} and Magnetic-Title~\cite{huang2020surface}. The experimental results demonstrate that our model has achieved state-of-the-art performance in the defect classification task.

\section{Related Work}
\subsection{Traditional Defect Inspection}
Traditional machine vision-based methods primarily include structural methods~\cite{mak2009fabric}, filter-based methods ~\cite{bai2014saliency}, and model-based methods ~\cite{wang2018simple}. These methods typically extract texture features and utilize support vector machines (SVM)  or back propagation (BP) neural networks for classification. Among the traditional methods, statistical methods consider texture as a random phenomenon and analyze the distribution of random variables to describe texture. Aghdam \textit{et al.} ~\cite{aghdam2012fast} proposed the local binary model, which is more robust to grayscale changes caused by illumination changes. Spectral methods are used to detect defects from specific domains, such as the frequency domain and spatial domain, and the Gabor filter is a classic spectral approach. Model-based methods, such as the Markov Random Field Model and Bayesian model, aim to detect defects by projecting them to a low-dimensional distribution space.

However, the design of features is complicated and relies on the experience of experts, which limits the large-scale application of conventional detection methods in industrial production. These methods require experts to perform complex and lengthy tuning and design, resulting in significant labor costs.

\subsection{Deep Learning-Based Defect Inspection}

Detection-based~\cite{yu2022net,zeng2022small,yang2022semiconductor} and segmentation-based~\cite{yu2022net,damacharla2021tlu,li2021defectnet} defect inspection methods have achieved some success. \cite{he2019end} is a detection-based method, it introduces a method to enhance the quality of defect proposals which integrated multi-level features into one feature and was subsequently input into the region proposal network to generate high-quality defect proposals. Chen \textit{et al}~\cite{chen2017automatic} cascaded two detectors (SDD and YOLO) and a classifier directly to check defects of fasteners. Mei \textit{et al}~\cite{mei2018unsupervised} proposed a segmentation method that first applied an encoder-decoder architecture to extract features, and then defined a segmentation threshold to locate the defect areas. The other segmentation-based method \cite{song2020edrnet} incorporates an attention mechanism into its architecture to steer it to focus more on defect features.
\begin{figure*}[htb]
	\centering
	\includegraphics[width=\textwidth]{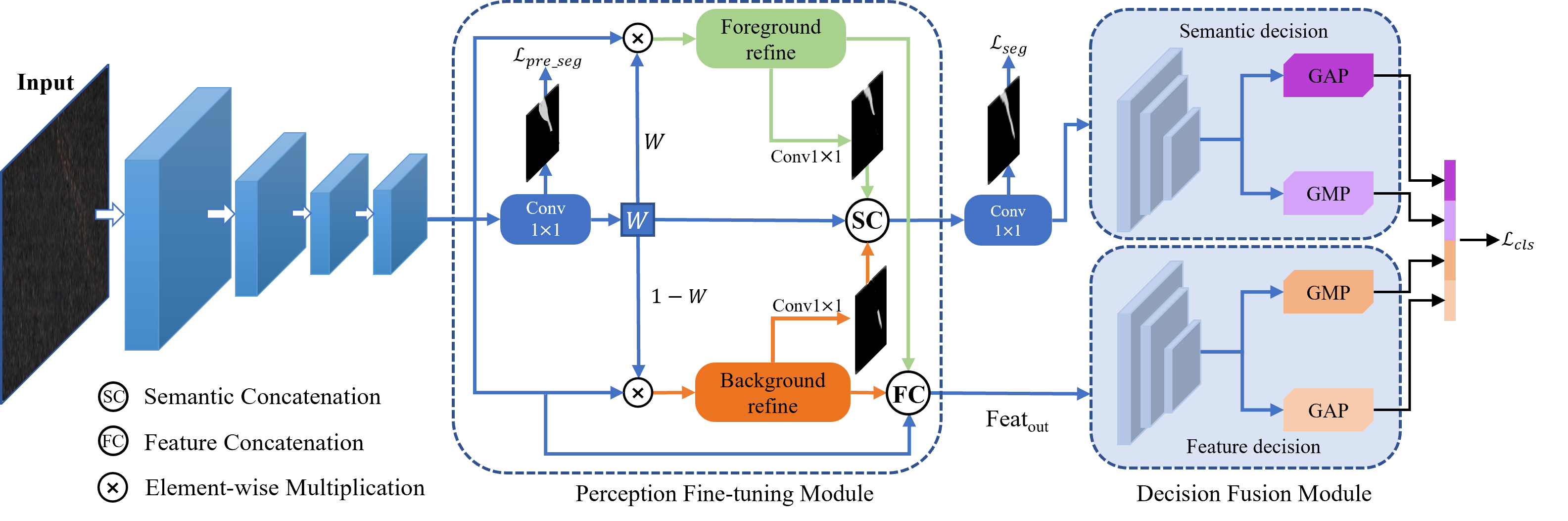}
	\caption{The overall network structure of the proposed method. It contains two stages including segmentation and classification. The segmentation stage consists of the feature extraction backbone and the perception fine-tuning module. The classification stage has a decision fusion module consisting of the semantic decision and feature decision.}
	\label{frame}
\end{figure*}

The discernment of image types holds significant importance within practical applications, thereby rendering the defect classification task as the central focus of our study. Racki \textit{et al.}~\cite{racki2018compact} leverages the local connectivity, shared weights, and abstract feature extraction capabilities of Convolutional Neural Networks (CNN) to endow the network with multi-level feature extraction capabilities for detecting defects. Xie \textit{et al}.~\cite{xie2022pyramid} have proposed a novel one-stage architecture known as the Pyramid Grafting Network (PGNet), which transfers global information from the transformer branch to the CNN branch at the decoder. Zhang \textit{et al}.~\cite{zhang2020mcnet} integrate multiple contextual information through the Pyramid Pooling Module (PPM) and attention module, aiming to enhance defect features and filter out noise. Dong~\cite{dong2020defect} \textit{et al}. employed genuine defects with diverse backgrounds to generate an extensive collection of synthetic images. These synthetic images were then utilized to train a classification network, addressing the issue of sparse labels. To harness the location information of defects and obtain more accurate classification results, Bo{\v{z}}i{\v{c}} \textit{et al}.~\cite{bovzivc2021end} have successfully introduced pixel-level labeling into image-level detection tasks by balancing the losses in different stages.

However, their models do not account for the low-contrast problem and background similarity problem, which makes it challenging to classify non-uniform areas and defects similar to the background. To this end, we propose a two-stage decision fusion classification network.
Different from existing methods, our method incorporates the semantic decision with the feature decision in the classification stage to strengthen the decision ability of the network. In the segmentation stage, the network separates the foreground and background and fine-tunes the corresponding features to enhance the ability to extract defect-related features.

\section{Proposed Method}

In this section, we present a decision fusion network with perception fine-tuning for defect classification, which is named DFNet. DFNet consists of the segmentation stage and the classification stage. The overall structure of DFNet is shown in Figure 2. The segmentation stage contains a feature extraction backbone and a perception fine-tuning module (PFM). The classification stage has a decision fusion module (DFM) to do the classification task. We mainly introduce PFM and DFM in this section. In addition, we introduce the inner-outer separation weight matrix (SWM) which is used in the segmentation loss calculation. 

\subsection{Perception Fine-tuning Module}
The feature extraction part of the segmentation stage is founded on the architecture of the first stage introduced in~\cite{bovzivc2021mixed}. We introduce a perception fine-tuning module (PFM) in the segmentation stage, which aims to refine the features and the segmentation results.

The PFM module begins by using a $ 1\times 1 $ single-channel convolution to generate the initial segmentation map and a weight map. This weight map is then used to separate the foreground features from the background features through the point product operation with the original feature map. Following this separation, the foreground refining and background refining modules are adopted to fine-tune the corresponding features.
Both the foreground and background refining modules involve the application of four 3$ \times$3 convolutional layers. It is noted that the last two layers of the refining modules employ dilated convolutions with expansion rates of 2 and 5, respectively. Its purpose is to exploit rich contextual information to refine the features and reduce the interference caused by the background. Finally, the refined foreground and background features are merged with the initial input features of PFM and sent to the classification stage.

Simultaneously, the refined foreground and background features are used to generate the corresponding complementary segmentation maps. These segmentation maps are combined with the initial segmentation map to produce the final segmentation result. The final segmentation result is then forwarded to the classification  stage.  As a result, the PFM module provides rich information including refined features and segmentation results for the final classification decision.

Furthermore, it is noted that both the initial segmentation and the final segmentation in the PFM module undergo supervision by the segmentation ground truth during the learning procedure. This supervision learning process helps the network to perceive more robust features and segmentation results.  

\subsection{Decision Fusion Module}
In the classification stage, to achieve accurate classification outcomes, it is important for the decision-making layer to incorporate the output results from the segmentation stage effectively. To this end, we propose a decision fusion module (DFM) that combines the semantic decision branch and the feature decision branch in the classification stage to improve the accuracy of classification. The process of the decision fusion model is shown in Figure 3. The right bottom part demonstrates the generation of the semantic branch and feature branch pertaining to decision-making, while the right upper part depicts the generation and fusion of semantic decision and feature decision.

The DFM consists of two branches: the semantic decision branch and the feature decision branch. The output feature of the segmentation network is used as the input for the feature decision branch, while the output semantic segmentation results of the segmentation network are used as the input for the semantic decision branch. 
In the feature decision branch, we compress the 1152 channel features into 32 channels using convolutional layers of sizes 5$\times$5$\times$8, 5$\times$5$\times$16, and 5$\times$5$\times$32, with each followed by a maximum pooling layer. We then perform a global maximum pooling layer and a global average pooling layer to obtain two vectors of volume 32$\times$1, which we refer to as A and B, respectively. 
For the semantic decision branch, in order to encode the one-channel map as a decision vector, we apply the same operations as the feature decision branch to generate the semantic maps. Subsequently, semantic decision vectors are generated through global max pooling and global average pooling, resulting in two vectors with dimensions of 32×1, denoted as C and D respectively. A fusion process uses the concatenation operation to combine A, B, C, and D, resulting in a 128×1 vector. This vector encompasses both semantic and feature information. Finally, the fully connected layer outputs the classification result.
\begin{figure}
	\centering
	\includegraphics[width = 0.48\textwidth]{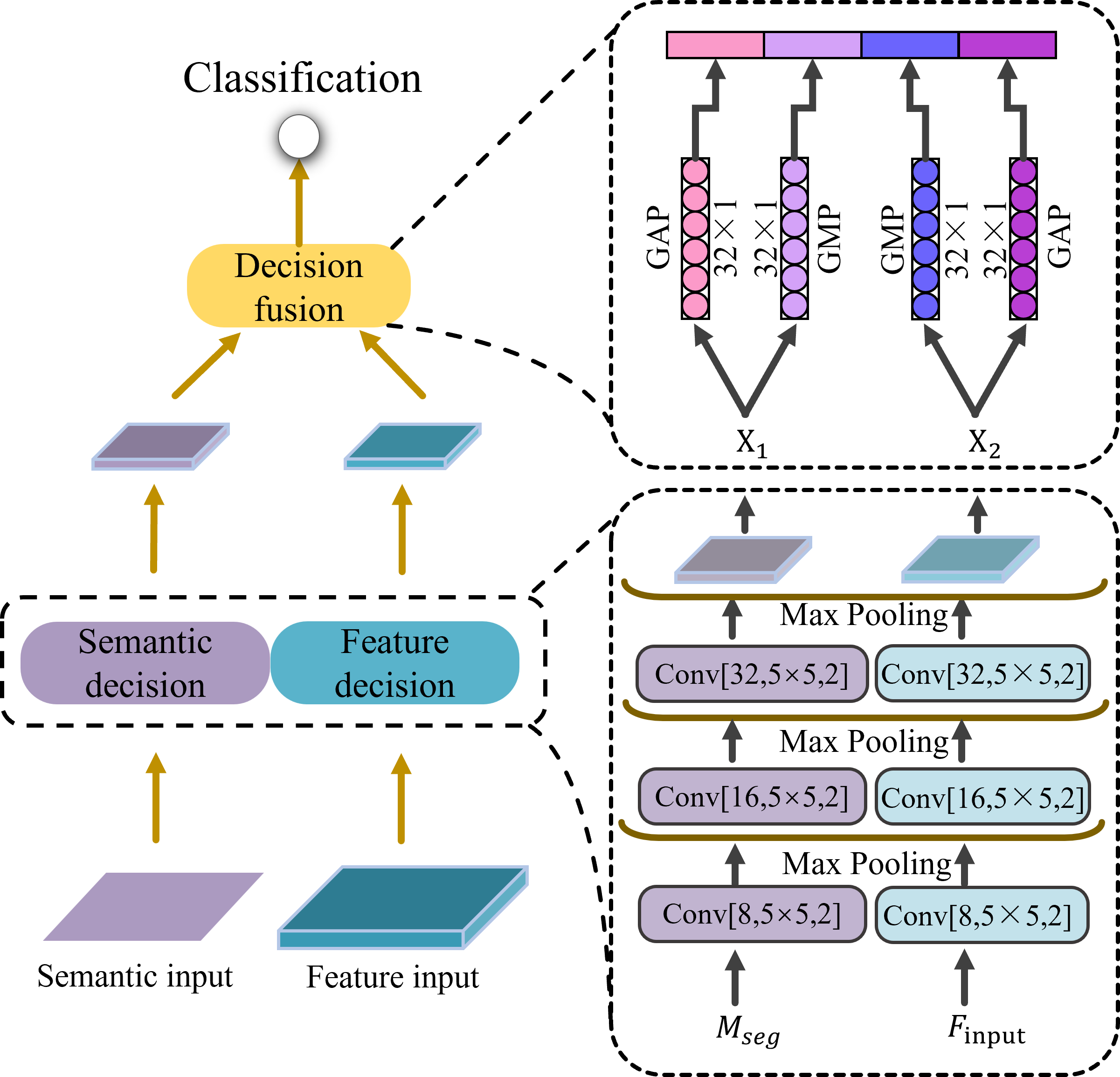}
	\caption{Architecture of DFM. Showing the details of semantic decision-making, feature decision-making, and decision fusion, respectively}
\end{figure}

%
%
As a result, the incorporation of both the semantic decision and feature decision contributes to the classification decision-making process, empowering our network to concentrate on the pertinent characteristics of the classification task.

\subsection{Loss function}
We employ an approach for loss calculation that effectively ensure the convergence of the model and improving training stability. This is achieved by combining the losses from both the segmentation and classification tasks. The total loss function is defined as Equation(3) where $n$ represents the current number of epochs,  $n_{e p}$ represents the total number of epochs,  ${L}_1$ represents the loss of classification and ${L}_2$ represents the loss of segmentation. This method prioritizes weight updates for the PFM and feature extraction network during the initial training stage, and focuses on weight updates for the DFM during the later stage, therefore mitigates the influence of inaccurate segmentation results on classification during the early training phase.
\begin{equation}
\mathcal{L}_{t o t a l}=\frac{n}{n_{e p}} \cdot \mathcal{L}_1+\left(1-\frac{n}{n_{e p}}\right) \cdot \mathcal{L}_2
\end{equation}

The weight matrix for the segmentation loss is denoted as  $M\left( {pix} \right)$. The loss $\mathcal{L}_{pre\_seg}$ corresponds to the semantic segmentation loss generated by feature extraction and $\mathcal{L}_{seg}$ corresponds to the semantic segmentation loss after foreground and background refinement, both of these losses are calculated from the binary cross-entropy loss. These two losses are combined with the classification loss $\mathcal{L}_{cla}$  which is calculated using the cross-entropy loss function. To address different classification tasks, we employ the parameter $\delta$, which takes the form of the Sigmoid function for binary classification tasks and the Softmax function for multi-classification tasks. The classification prediction generated by the fully connected layer is denoted as $p$.
\begin{equation}
\mathcal{L}_1=M( pix ) \cdot \left(\mathcal{L}_{pre\_seg}+\mathcal{L}_{seg}\right)
\end{equation}
\begin{equation}
\mathcal{L}_{2} = \mathcal{L}_{cla}\left( \delta(p) \right)
\end{equation}

This approach for loss calculation involves initializing the coefficient of the segmentation loss close to 1 and the coefficient of the classification loss close to 0 at the start of training. This initial configuration allows for rapid updates of the segmentation network, minimizing its impact on the classification network. As training progresses, the coefficient of the classification loss becomes larger.

\subsection{Separation Weight Matrix}
Data augmentation is a commonly used technique in machine learning. One such technique is dilation-based data augmentation, which involves expanding the area of annotated defective regions to allow the model to focus on the surrounding contextual information. This method increases the number of defective pixels in the annotation, thereby balancing the number of positive and negative samples. Experiments conducted in ~\cite{tabernik2020segmentation} demonstrate that this method can effectively improve classification accuracy. However, this technique can also lead to a reduction in the accuracy of segmentation results as the dilation factor increases. To address this issue, an edge-center distance attenuation weight matrix is often employed to reduce the weight of edge losses. Specifically, the weight is inversely proportional to the distance from the pixel to the edge. However, this attenuation method presents a significant problem as it also reduces the weight of accurately annotated regions. 
	  \begin{figure}[t]
		\centering
		\includegraphics[width=0.48\textwidth]{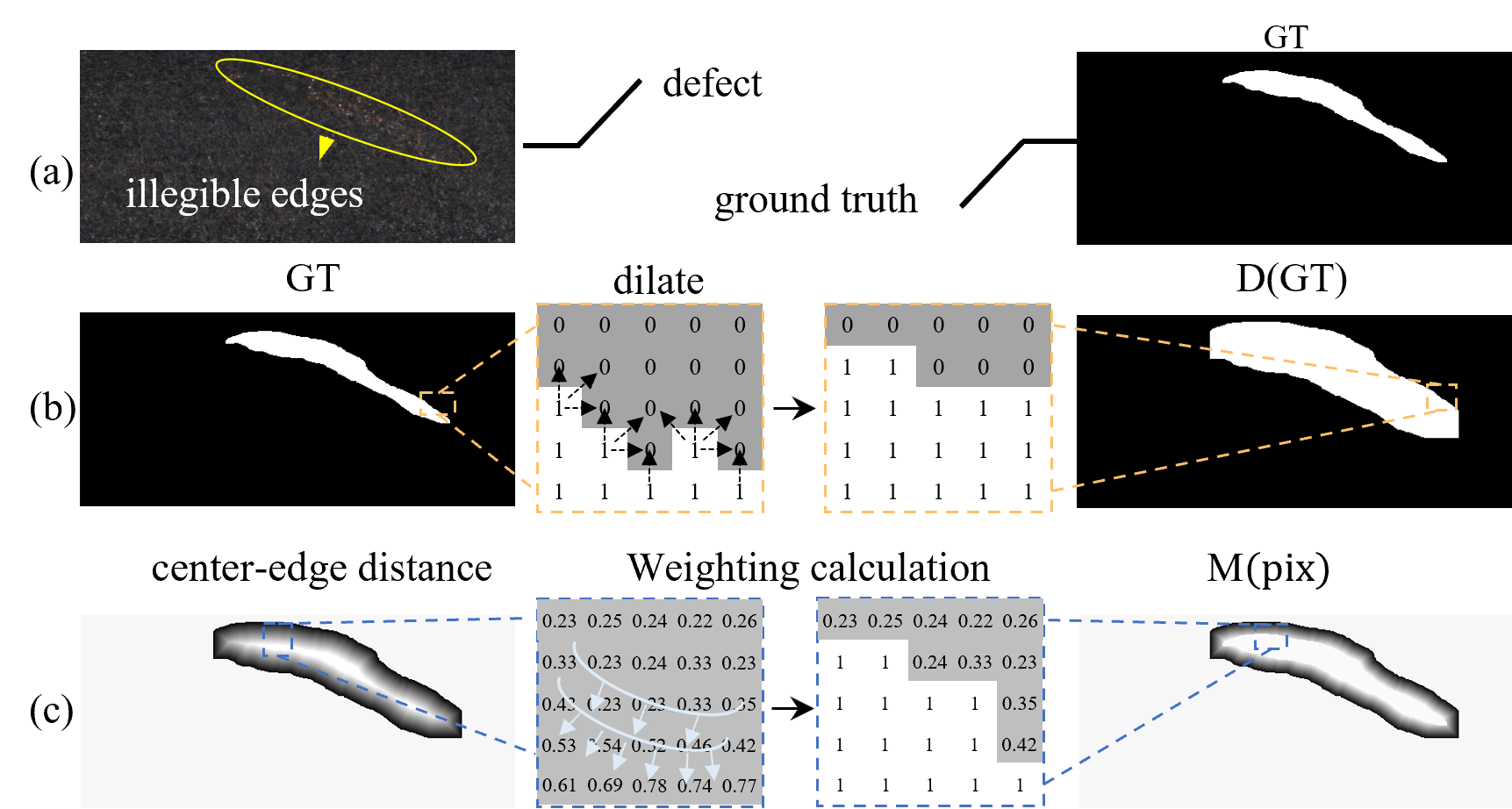}
	\caption{The generation of the separation weight matrix. (a) denotes the defective image and the GT image, (b) denotes the GT image after the dilation operation, and (c) denotes the weight matrix generated by the separation weight transformation operation.}
		\label{frame}
		\end{figure}
  
To overcome this issue, we introduce a separation weight matrix (SWM) that resets the weight of accurately annotated regions to 1 and only applies weight attenuation to the dilated regions. As shown in equation (1) and (2). This method effectively reduces the segmentation weight assigned to the outer side of the defect, thereby minimizing the potential for misjudgment at the edge. Additionally, this approach leverages the benefits of pixel expansion data enhancement, which enhances the confidence level in identifying the inner side of the defect.

\begin{equation}
M\left( {pix} \right) = \left\{ \begin{matrix}
{\frac{D(pix)}{D_{max}\left( {pix} \right)},~~D\left( {pix} \right) \leq d} \\
{1~~~~~~~~~~~,~~D\left( {pix} \right) > d} \\
\end{matrix} \right.~
\end{equation}
\begin{equation}
\mathcal{L}_{seg}\left( {pix} \right) =  {M\left( {pix} \right)} \cdot \hat{\mathcal{L}}(pix)
\end{equation}
 $\hat{\mathcal{L}}(pix)$ denotes the pixels of the original defect loss. $D(pix)$ is the distance transformation function, which yields the Euclidean distance between the points in the defect region of GT and the most recent pixel with a value of 0. $D_{max}({pix})$ represents the maximum distance. $M\left(pix\right)$ is the weight matrix of GT labels generated by distance transformation. An example of a split mask and weight is shown in Figure 4(c). The weight of defect-free pixels in the segmentation mask is 1.

\section{Experiments}

\subsection{Datasets}

We evaluated the proposed network using two publicly available datasets: KolektorSDD2~\cite{bovzivc2021mixed} and Magnet-Tile-defect-dataset~\cite{huang2020surface}.

\textbf{KolektorSDD2.} It contains 3335 images of electrical commutators. The images have a width of approximately 230 pixels and a height of around 630 pixels. The training set consists of 246 defective positive cases (defect images) and 2085 defective negative cases (non-defect images). The test set contains 110 defective positive cases and 894 defective negative cases. The defects in this dataset are annotated at the pixel level and include scratches, spots, and large surface defects. Notably, this dataset presents a significant challenge due to the presence of small-scale defects with blurred edges, which greatly affects the network's identification capability.

\textbf{Magnetic-Tile-defect-dataset.} It comprises six categories of magnetic tile surfaces, named Blowhole, Break, Crack, Fray, Uneven, and Free. The dataset contains a total of 1344 images. To create the training and testing sets, we randomly divided each category in a 7:3 ratio and then combined the training sets of each category. Similarly, the test sets of each category were combined to form the final test set. Consequently, we obtained 940 training examples and 404 testing examples. Given the significant variation in picture sizes, all the images were adjusted to a resolution of 512$\times$512 pixels. This dataset exhibits diverse features, encompassing intricate and random lighting conditions, as well as a wide range of defects and backgrounds. Furthermore, the presence of numerous uneven backgrounds that resemble defects further complicates the accurate recognition of defects.

\subsection{Implementation Details}
Extensive experiments were conducted on a machine equipped with a 3060Ti GPU (8GB) and an Intel(R) i7-10700K CPU, utilizing the PyTorch library. We will support it with MindSpore.

Our model was trained using the Adam optimizer. To ensure consistent end-to-end training, we implemented a strategy for truncating gradient propagation from the classification network to the segmentation network during the initial stages of training. By constraining the flow of gradients, we prevented the classification network from exerting excessive influence on the segmentation network. This approach allows for independent updates of both stages. To address the issue of imbalanced positive and negative samples, we alternated between inputting positive samples and negative samples during the iterations. Once all positive samples were utilized once, we considered one training epoch to be completed.

In our study on KolektorSDD2, we set the number of training epochs to 100, employing a learning rate of 0.01 and a batch size of 1. Since the dataset contains numerous minute defects, we discovered that using larger morphological expansion sizes yielded improved classification results. Therefore, we employed a 25$\times$25 kernel for the dilated augmentation on annotations. Considering the complex and challenging nature of the Magnetic-Title dataset, we extended the training epoch of the classification network to 200, with a learning rate of 0.001 and a batch size of 1. Additionally, given the prevalence of large defect shapes within this dataset, we utilized a 7$\times$7 kernel for the dilated augmentation on annotations.

\subsection{Metrics}

In defect classification, the number of defect samples is often significantly smaller than the number of normal samples, highlighting the importance of accurately evaluating the sample imbalance. This paper employs the average precision (AP) and mean average precision (mAP) as performance measures. Average precision is determined by calculating the area under the precision-recall curve, allowing for a precise assessment of the model's effectiveness when dealing with datasets that exhibit substantial differences between positive and negative samples. In order to provide a detailed analysis of the model's classification performance on positive and negative images, True Positive Rate (TPR) and True Negative Rate (TNR) are also utilized as evaluation metrics. The calculation of these evaluation metrics is depicted in Equations 6 and 7. Furthermore, it employs the following abbreviations in the experimental results: TP (True Positive), FN (False Negative), TN (True Negative), and FP (False Positive), representing the different classifications of the results.
\begin{equation}
TPR=\frac{TP}{TP + FN}, TNR=\frac{TN}{TN + FP}
\end{equation}
\begin{equation}
Accuracy~ = ~~\frac{TP + TN}{TP + FP + TN + FN}.
\end{equation}
\subsection{Experimental Results}

We compare our method with other state-of-the-art classification models. Among them, GoogLeNet~\cite{szegedy2015going}, 
		     \begin{figure}[!h]
		\centering
	\includegraphics[width=0.48\textwidth]{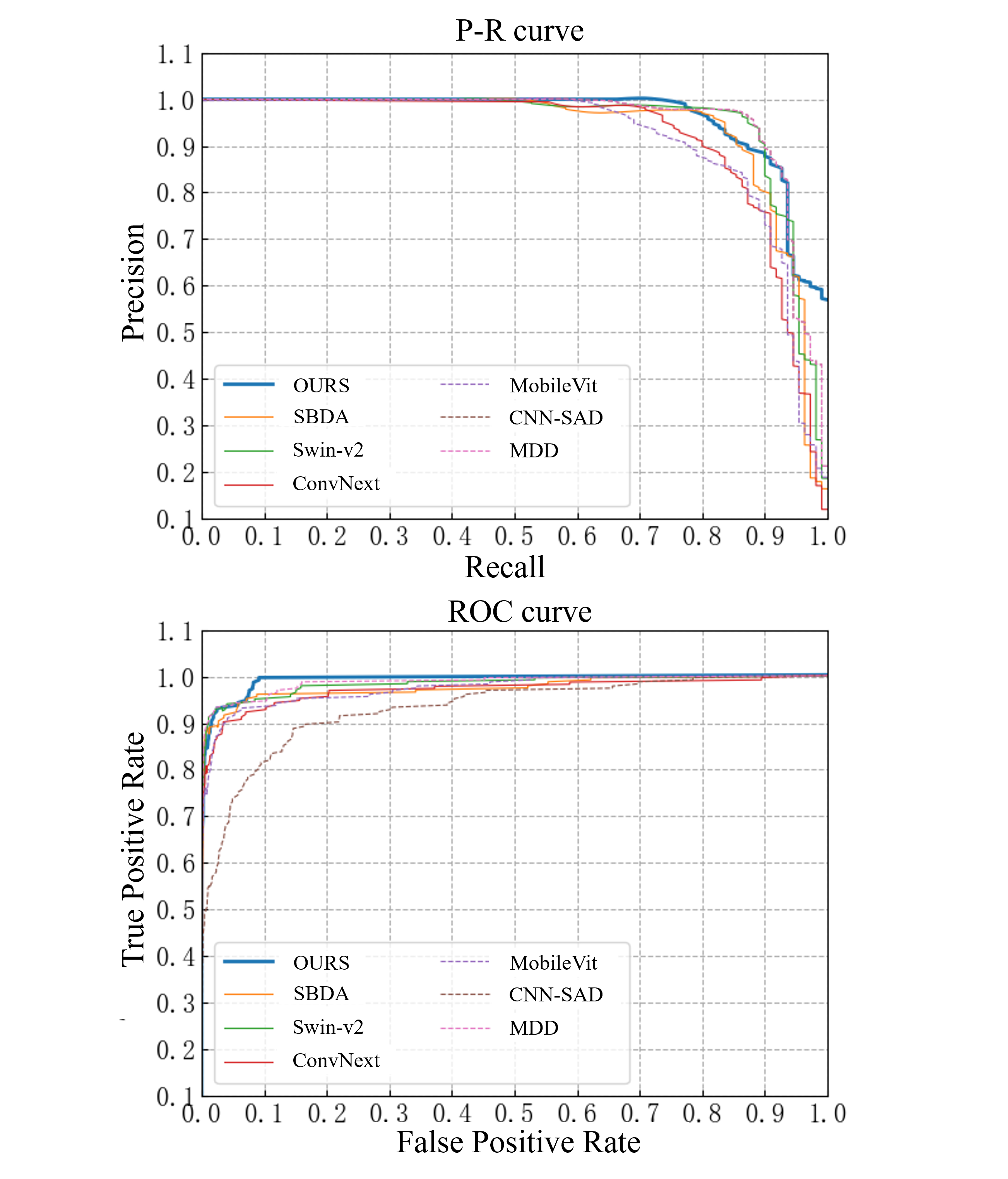}
		\caption{Comparison of P-R curves and ROC curves of different methods on KolektorSDD2.
		}
		\label{frame}
		\end{figure}	
MobileVit~\cite{mehta2021mobilevit}, EfficientNetV2~\cite{tan2021efficientnetv2}, ConvNet~\cite{liu2022convnet}, Swin-v2~\cite{liu2022swin}, and SDDNet~\cite{cui2021sddnet} are single-stage networks that only use image-level label supervision.
CNN-SAD~\cite{racki2018compact}, SBDA~\cite{tabernik2020segmentation}, MDD~\cite{bovzivc2021mixed} are two-stage networks that use both pixel-level labels and image-level label supervision. 

   \begin{table}[t]
         \large
         \tabcolsep=0.15cm
			  \resizebox{0.49\textwidth}{!}{
		\begin{tabular}{c|c|cccccc}
		\hline
		Method               & AP(\%) & TPR      & TNR    & Accuracy & TP/FN & TN/FP   \\ \hline
		GoogLeNet          & 90.1      & 80.9    & 99.2     & 97.2  & 89/21  & 887/7  \\ 
		MobileViT           & 93.2      & 82.7    & 99.3     & 97.5  & 91/19  & 888/6  \\ 
		EfficientNetV2        & 94.2      & 86.3    & 99.2     & 97.8  & 95/15  & 887/7  \\ 
		ConvNet          & 93.5      & 83.6    & 99.5     & 98.2  & 92/18  & 890/4  \\ 
		Swin-v2          & 94.9      & 82.7    & 99.5     & 97.9  & 93/17  & 890/4  \\ \hline		
		CNN-SAD          &93.2      & 87.2    & 98.7     & 97.5  & 96/14  & 883/11  \\ 
		SBDA	  &94.6	 &86.3	 &99.1	&97.7	 &95/15	&886/8\\ 
		MDD          & 95.4 &86.3      & 99.4    & 98.0     & 95/15  & 889/5  \\ \hline		
		OURS           & \textbf{96.1} & \textbf{89.0} & \textbf{99.4}  & \textbf{98.3}  & \textbf{98/12} &\textbf{889/5} \\ \hline
		\end{tabular}
			  }
		\caption{Quantitative comparison of various methods on the KolektorSDD2 dataset, encompassing specialized defect detection networks as well as existing classification networks.}
		\label{tab:booktabs}
			  \end{table}

\textbf{Results on KolektorSDD2.} The outcomes are showcased in Table 1, while Figure 5 exhibits the P-R curve and ROC curve. This table includes the classification results of other networks as well. Due to the challenging nature of the KolektorSDD2 dataset, many state-of-the-art classification networks struggle to achieve satisfactory performance. Our proposed method achieves an improvement of 1.2 points (from 94.9\% to 96.1\%) compared to the existing single-stage classification network Swin-v2. Additionally, it outperforms the existing two-stage network by 0.7 percentage points (from 95.4\% to 96.1\%). These results indicate that our method attains state-of-the-art performance under fully supervised conditions.

\textbf{Results on Magnetic-Title dataset.} To further assess the effectiveness of the proposed method, experiments were conducted on the Magnetic-Tile dataset. Table 2 presents the classification results obtained in the experiments. The experimental results demonstrate that our proposed network exhibits classification capability for defects of varying scales and shapes in the dataset. It also shows that it can achieve outstanding performance in multi-classification tasks. The mean average precision (mAP) result reaches 94.6\%, surpassing the performance of both existing one-stage and two-stage networks.

    \begin{table}[t]
        \large
        \tabcolsep=0.10cm
	  \resizebox{0.49\textwidth}{!}{
		\begin{tabular}{c|c|ccccccc}
		\hline
		Method                 & mAP(\%) & Blowhole & Break & Crack & Fray & Uneven & Free  \\ \hline 
        GoogLeNet & 91.1    & 100.0    & 95.6  & 92.8  & 81.8 & 76.4   & 99.9  \\
		MobileVit          & 92.0    & 96.4    & 94.3  & 91.2  & 93.4 & 77.2   & 98.9  \\
		ConvNet          & 92.2    & 95.6    & 90.2  & 93.5  & 93.9 & 80.6   & 99.9  \\ 
		Swin-v2         & 93.0    & 92.5    & 95.3  & 95.1  & 94.1 & 81.3   & 99.7  \\ 
		SDDNet    & 93.6    & 98.7 & 95.3    & 100.0  & 83.9  & 83.8 & 99.9  \\ \hline
		CNN-SAD          & 91.1    & 90.5     & 93.2  & 100.0 & 86.4 & 76.8  & 99.9  \\ 
		SBDA	   & 91.4    & 92.6     & 90.3  & 100.0 & 83.6 & 82.2  & 99.9  \\ 
		MDD      & 93.8    & 96.8     & 86.6  & 100.0  & 90.0 & 91.2   &   98.8 \\  \hline		
		OURS                 &\textbf{94.6}    & \textbf{94.8}     & \textbf{89.9}  & \textbf{100.0}  & \textbf{92.3} & \textbf{91.7}   &   \textbf{98.9}    \\ \hline
		\end{tabular}
	  }
	  \caption{Comparison of mAP and AP of diverse surface defects on Magnetic-Tile dataset. }
	  \label{tab:booktabs}
	  \end{table}


\section{Ablation Study}

   \begin{table}[t]
        \large
        \tabcolsep=0.25cm
	  \resizebox{0.48\textwidth}{!}{
		\begin{tabular}{c c c|cc|cc}
		\hline
		 PFM & DFM & SWM &\multicolumn{2}{c|}{KolektorSDD2} &\multicolumn{2}{c}{Magnetic-Tile}\\ \hline
	- & - & - & AP   & FP+FN & mAP & FP+FN   \\ \hline
	- & - & - & 93.3 &6+21   & 91.9  &9+14        \\ \hline
	\checkmark & - & - & 94.1 &8+14   & 92.3  &8+13 \\ \hline
		 \checkmark & \checkmark & - & 95.2 &5+15   & 93.1 &4+12        \\ \hline
		\checkmark & \checkmark & \checkmark &96.1 &5+12   & 94.6 &3+8\\ \hline
		\end{tabular}
	  }
	\caption{Ablation study on the proposed modules on two datasets, reporting the average precision (AP), the number of false positive (FP) examples, and the number of false negative (FN) examples.}
	  \label{tab:booktabs}
	  \end{table}

In this study, we investigate the impact of individual components in the proposed model: the perception fine-tuning module, the decision fusion module, and the separation weight matrix. Ablation studies were conducted on KolektorSDD2 and Magnetic-Tile datasets, using the same parameter settings mentioned above during training. The performance was evaluated by progressively enabling and disabling individual components while keeping the remaining components constant. The results are presented in Table 3, which demonstrates that the model's performance improved on both datasets when all three components were incorporated. Specifically, the average precision (AP) results increased by 2.8 percentage points (from 93.3\% to 96.1\%) on KolektorSDD2 and by 2.7 percentage points (from 91.9\% to 94.6\%) on Magnetic-Tile. These findings highlight the significance of each component in enhancing the overall model performance. In the following sections, we provide detailed descriptions of the contribution of each component to the observed improvements in the results.
   
\textbf{Perception Fine-tuning Module.} In the context of defect inspection, the segmentation task often encounters challenges in accurately identifying defects due to their small sizes and the significant variability in foreground and background elements. The PFM method aims to tackle these challenges by emphasizing the features of foreground defects and suppressing interference of the background. 
 
The experimental results demonstrate that PFM effectively enhances the performance of the network. On the KolektorSDD2 dataset, the average precision (AP) increases by 1.9 points (from 93.3\% to 95.2\%). On the Magnetic-Tile dataset, it improves the mean average precision (mAP) by 0.4 points (from 91.9\% to 92.3\%). 

The impact of incorporating PFM can be observed in Figure 6, where the heat map demonstrates the improved perception of features of the defective regions by the network. This enhancement leads to an increase in the confidence level of foreground defects. Moreover, the introduction of PFM can help filter out background interference. Overall, the integration of the PFM module effectively enhances the defective regions and suppresses background interference in the heat map, as depicted in the provided examples. These results demonstrate the effectiveness of the proposed PFM in identifying defects with complex backgrounds.

\begin{figure}
	\centering
	\includegraphics[width = 0.48\textwidth]{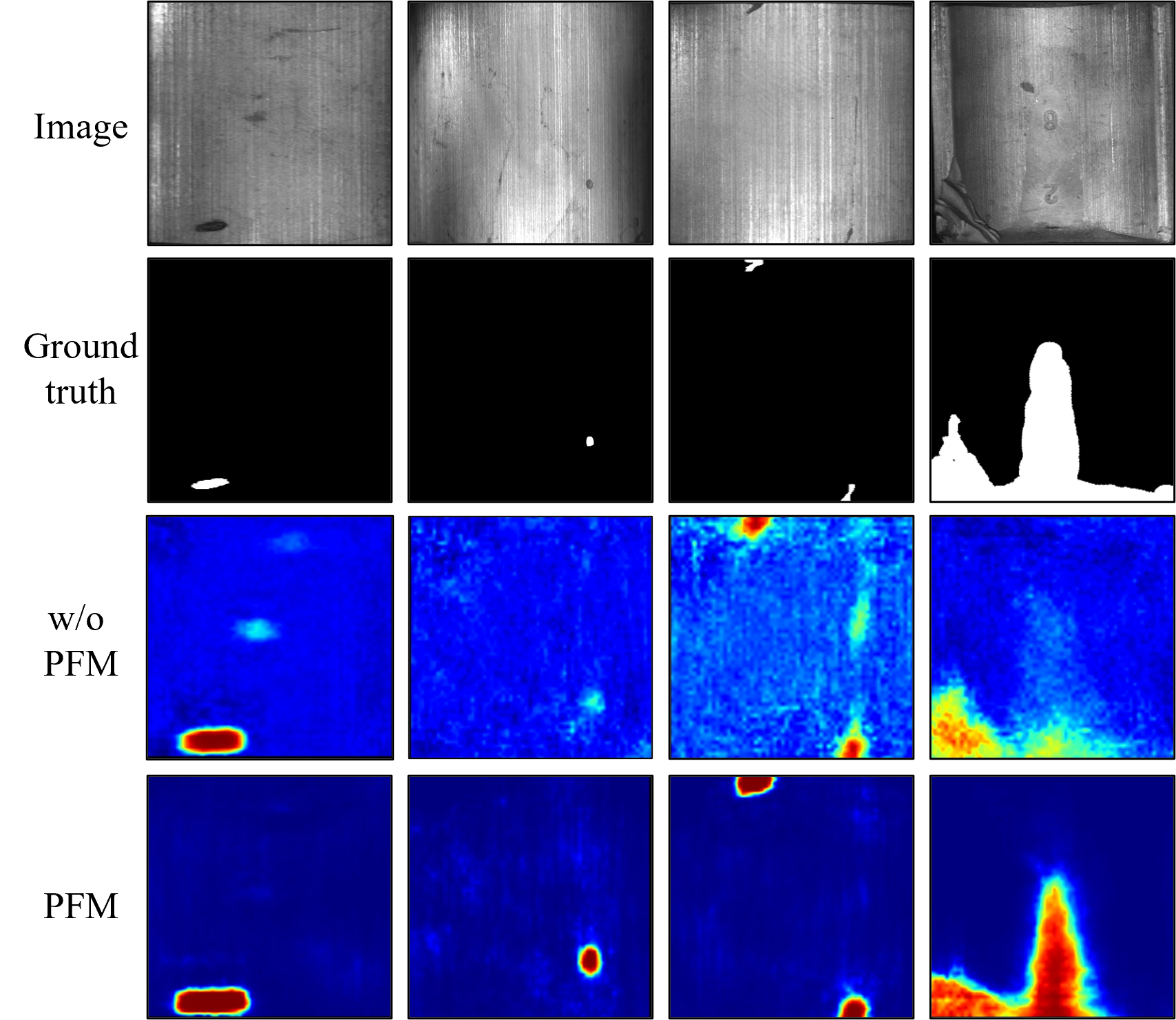}
	\caption{The first row displays the image examples that contain defects with different appearances from the Magnetic-Title dataset. The second row exhibits their corresponding ground-truth images. The third and fourth rows present the comparison of the heat maps generated without PFM and with PFM, respectively.}
\end{figure}

\textbf{Decision Fusion Module.} It is seen from Table 3 that the combined utilization of PFM and DFM obtains an improvement in AP compared to the individual utilization of PFM on both datasets. On the KSDD2 dataset, there is an increase of 1.1 points (from 94.1\% to 95.2\%). On the MT dataset, there is an improvement of 0.9 points (from 92.3\% to 93.1\%). Additionally, the incorporation of DFM leads to a reduction in false positives, indicating its efficacy in suppressing background interference. These results affirm the effectiveness of the proposed DFM module.

\textbf{Separation Weight Matrix.} The ablation study indicates that the incorporation of the separation weight matrix (SWM) enhances the defect classification ability. The evaluation on the KolektorSDD2 dataset reveals an increase of 0.9 points (from 95.2\% to 96.1\%) in average precision (AP). Similarly, on the Magnetic-Tile dataset, the employment of SWM results in an improvement in mean average precision (mAP) by 1.5 points (from 93.1\% to 94.6\%). Moreover, the utilization of the separation weight matrix effectively mitigates the wrong classifications for both positive and negative samples, resulting in a reduction of the number of false positives (FP) and false negatives (FN) by 3 and 5 in the KolektorSDD2 and Magnetic-Tile datasets, respectively. These outcomes substantiate the effectiveness of the proposed SWM.

\section{Conclusion}
In this paper, we propose a decision fusion network (DFNet) for the defect classification task. DFNet consists of a segmentation stage and a classification stage and is learned in an end-to-end way with pixel-level labels and image-level labels. 
DFNet owns a perception fine-tuning module (PFM) in the segmentation stage and a decision fusion module (DFM) in the classification stage. 
The PFM refines the network features and segmentation results by first focusing on the foreground and the background regions separately and then converging all the information. As a result, PFM enhances the defect features and suppresses the background interference. 
The DFM incorporates the output features and segmentation results of the PFM module and generates the semantic decision vector and feature decision vector from them, respectively. It then converges the decision vector and makes a final decision.
Furthermore, the separation weight matrix is introduced to tackle the issue of pixel-level label dilation. It highlights the weights on the inner side of defects during the learning procedure and thus enhances the ability to detect small defects.
The experimental results on two publicly available industrial datasets demonstrate the superiority of our method compared to other existing methods.

\section{Acknowledgments}
This work was supported in part by the Nation Key Research and Development Program of China under Grant 2021YFB3301500; in part by the National Natural Science Foundation of China under Grant 62172371, U21B2037, 62102370, 62202433; in part by Natural Science Foundation of Henan Province under Grant 232300421093; in part by the Foundation for University Key Research of Henan Province under Grant 21A520040 and Key R\&D and Promotion Projects of Henan Province under Grant 222102210126; in part by CAAI-Huawei MindSpore OpenFund.

\bibliographystyle{named}
\bibliography{DFNet}

\end{document}